\documentclass[a4paper, 10pt, conference]{ieeeconf}      
\usepackage{FG2024}

\FGfinalcopy 
\usepackage{listings}
\usepackage{algorithm}
\usepackage{algpseudocode}
\usepackage{pifont}

\usepackage{graphicx}
\usepackage{float}
\usepackage{adjustbox} 
\usepackage{multirow}
\usepackage{amsmath}
\usepackage{threeparttable}
\usepackage{balance}

\IEEEoverridecommandlockouts                              
\def\FGPaperID{****} 

\title{\LARGE \bf
 Sum of Group Error Differences: A Critical Examination of Bias Evaluation  in Biometric Verification and a Dual-Metric Measure  
}

\author{\parbox{16cm}{\centering
    {\large Alaa Elobaid $^{1,2}$,  Nathan Ramoly $^{3}$,  Lara Younes $^{3}$,  Symeon Papadopoulos $^1$,  Eirini Ntoutsi $^4$, and Ioannis Kompatsiaris$^1$}\\
    {\normalsize
    $^1$ Information Technologies Institute, CERTH, Thessaloniki, Greece\\
    $^2$ Freie Universit\"at Berlin, Berlin, Germany\\
    $^3$ IDnow, Research CoE, Cesson-Sévigné, France\\
    $^4$ Research Institute CODE, Bundeswehr University Munich, Munich, Germany}\\
    Email: alaa.elobaid@fu-berlin.de}
}

\usepackage{fancyhdr}
\usepackage{hyperref}
\begin{document}

\ifFGfinal
\thispagestyle{empty}
\pagestyle{empty}
\else
\author{Anonymous FG2024 submission\\ Paper ID \FGPaperID \\}
\pagestyle{plain}
\fi
\maketitle

\thispagestyle{fancy}

\begin{abstract}
Biometric Verification (BV) systems often exhibit accuracy disparities across different demographic groups, leading to biases in BV applications. Assessing and quantifying these biases is essential for ensuring the fairness of BV systems. However, existing 
bias evaluation metrics in BV have limitations, such as focusing exclusively on match or non-match error rates, overlooking bias on demographic groups with performance levels falling between the best and worst performance levels, and neglecting the magnitude of the bias present. 

This paper presents an in-depth analysis of the limitations of current bias evaluation metrics in BV and, through experimental analysis, demonstrates their contextual suitability, merits, and limitations. Additionally, it introduces a novel general-purpose bias evaluation measure for BV, the ``Sum of Group Error Differences (SED$_G$)''. Our experimental results on controlled synthetic datasets demonstrate the effectiveness of demographic bias quantification when using existing metrics and our own proposed measure. We discuss the applicability of the bias evaluation metrics in a set of simulated demographic bias scenarios and provide scenario-based metric recommendations. Our code is publicly available under \href{https://github.com/alaaobeid/SEDG}{https://github.com/alaaobeid/SEDG}.
\end{abstract}

\section{Introduction}
Biometric Verification (BV) systems suffer from demographic biases that manifest in the form of different accuracy levels influenced by demographic factors such as skin colour, age, and gender \cite{grother_face_2019,cook_demographic_2019,albiero_face_2023}. These biases are transferred to real-world BV applications, such as unconstrained face verification \cite{maze_iarpa_2018} in mobile phone access \cite{lopez-lopez_dataset_2019}, biometric authentication \cite{bhattacharyya2009biometric} in identity-to-selfie matching \cite{shi2018docface}, and identity verification \cite{schroff2015facenet} in border control \cite{garcia_harms_2019}. However, there is no agreement over which metrics to use to quantify the amount of bias in a BV system \cite{serna_sensitive_2022}. Nevertheless, current bias evaluation methods and metrics suffer from at least one limitation at a time. Differential performance methods such as the fairness index measures \cite{101007}, and the analysis of Receiver Operator Characteristics (ROC), False Match Rate (FMR), and False Non-Match Rate (FNMR) curves \cite{lu_experimental_2019,krishnapriya_2019} do not reflect real-world scenarios where a decision threshold is required. Methods that do not account for global performance and cross-demographic impostors such as the genuine and impostor distributions \cite{krishnapriya_2019}, the demographic group verification accuracies \cite{gong_jointly_nodate,rfw}, the Standard Deviation (STD) in Group Equal Error Rates (EER$_G$) \cite{serna_sensitive_2022}, and the Gini Aggregation Rate for Biometric Equitability (GARBE) \cite{howard_evaluating_2023}, lack a global reference point for measuring bias and neglect the scenario where the impostor does not share demographic attributes with the genuine identity. The absence of a reference point for bias measurement can make it impossible to know the magnitude of the bias. Metrics that quantify bias in the form of maximum and minimum error rates, such as the Skewed Error Rate (SER) \cite{villalobos_fair_2022}, the Inequity
Rate (IR) \cite{howard_evaluating_2023} and the Fairness Discrepancy Ratio (FDR) \cite{de2021fairness} disregard demographic groups that exhibit
intermediate error rates. Hence, a system biased against one group is given the same score as a system biased against multiple groups as long as their minimum and maximum error values are identical. Additionally, most of these metrics require that an operational scenario is defined beforehand and may not be suitable for studying biases in systems in a manner that is independent of application.

This paper aims to highlight the limitations of current bias evaluation metrics in BV, evaluate their suitability and merits, and introduce a general-purpose bias evaluation measure that overcomes the identified limitations. The proposed bias evaluation measure captures deviations of demographic group error rates from global error rates, treats false matches and non-matches equally, and considers all demographic groups while calculating errors and when setting a suitable threshold. Our main contributions are as follows: 

\begin{itemize}
\item  We provide an analysis of bias evaluation metrics in the BV literature and identify their limitations.
\item We evaluate the effectiveness and behaviour of the bias evaluation metrics through experimental analysis in a range of simulated scenarios with different levels and types of bias.
\item We provide a detailed analysis of the strengths, weaknesses, and appropriateness of each metric in different bias scenarios.
\item We propose a novel dual-metric bias evaluation measure that overcomes the limitations of existing bias evaluation metrics in BV.
\end{itemize}


\section{Background and Related Work}
We first introduce the basics of generic Biometric Verification (BV) before discussing related research on bias and discrimination in BV.

\subsection{Biometric Verification Problem Formulation}
The generic BV problem is an application-independent formulation of various BV tasks that use an individual's biological traits to verify their identity. Therefore, it applies to various biometric data types, such as face images, fingerprints, speech, and iris data. BV is typically considered as a binary classification task, where the goal is to determine whether a pair of inputs is a \textbf{genuine pair}, meaning that the input pair genuinely belongs to the same person, or an \textbf{impostor pair} meaning that the pair belongs to different persons. In some cases, impostor pairs are selected to share the same demographic attributes. These are referred to as Within-Demographic Impostors (WDI) in contrast to Cross-Demographic Impostors (CDI), which do not necessarily share any demographic attributes.

More formally, the BV problem can be expressed as follows: Given a pair of biometric data samples $s_{1}$ and $s_{2}$, a biometric feature extraction model $M$ is used to extract two embedding vectors $e_1$ and $e_2$, which represent these samples and the distance between the pair of embedding vectors in our case the Euclidean Distance (ED) is then calculated as follows:
\[
e_1 = M(s_1)\]
\[
e_2 = M(s_2)\]
\[
\text{ED }(e_1, e_2) = \sqrt{\sum_{i=1}^{n} (e_{1i} - e_{2i})^2}
\]

The goal of the generic BV problem is to determine whether the pair of samples $s_{1}$ and $s_{2}$, which are being represented by the embedding vectors ($e_1$, $e_2$), belong to the same individual or not. It can be expressed as follows:
\[
D = \begin{cases}
    \text{True,} & \text{if ED ($e_1$, $e_2$)} \text{ is less than or equal to } T. \\
    \text{False,} & \text{otherwise.}
\end{cases}
\]
where:
\begin{itemize}
    \item $D$ is the decision indicating whether the pair of sample embeddings $(e_1, e_2)$ belong to the same individual or not.
    \item ED $(e_1, e_2)$ is the Euclidian distance between the pair of biometric sample embedding vectors e1 and e2.
    \item $T$ is the decision threshold used for verification.
\end{itemize}

In this problem, errors are usually quantified in terms of False Match Rates (FMR) and False Non-Match Rates (FNMR), which can be formulated as follows:

\begin{equation} \label{eq:FMR}
\mathrm{FMR} = \frac{\text{Number of false matches}}{\text{Total number of impostor pairs}}
\end{equation}
\begin{equation} \label{eq:FNMR}
\mathrm{FNMR} = \frac{\text{Number of false non-matches}}{\text{Total Number of genuine pairs}}
\end{equation}
where:
\begin{itemize}
    \item False matches are impostor pairs falsely flagged as genuine pairs.
    \item False non-matches are genuine pairs falsely flagged as impostor pairs.
\end{itemize}

Here, we define the True Match Rate (TMR) and the True Non-Match Rate (TNMR) as they are both used in our dataset generation process described in Section \ref{sec:dembias}.

\begin{equation} \label{eq:TMR}
\mathrm{TMR} = \frac{\text{Number of true matches}}{\text{Total Number of genuine pairs}}
\end{equation}
where:
\begin{itemize}
    \item True matches are genuine pairs correctly flagged as genuine pairs.
\end{itemize}
TMR can also be expressed  in terms of FNMR as:
\begin{equation} \label{eq:TMR_FNMR} \mathrm{TMR} = 1 - \mathrm{FNMR} \end{equation}
Similarly, the TNMR can be expressed  in terms of FMR as:
\begin{equation} \label{eq:TNMR_FMR} \mathrm{TNMR} = 1 - \mathrm{FMR} \end{equation}

\subsection{Bias and Discrimination in Biometric Verification} \label{sec:lit}
Bias in BV can be studied independently of any decision thresholds, referred to as Differential Performance, or by studying error rates at a decision threshold with a desired performance, referred to as Differential Outcomes. Both of those terms were introduced by \cite{howard2019effect}. Bias evaluation in terms of differential performance studies the differences in ROC curves or genuine-impostor distributions of the different demographic groups \cite{lu_experimental_2019, krishnapriya_2019,101007} independent of  application. However, the practicality of current differential performance approaches is questionable because, in real-world operational scenarios, a threshold needs to be determined in advance. Taking this into consideration, differential outcome methods are, by design, more representative of real-world verification scenarios and, therefore, are the focus of our paper.   

In the remaining part of this section, we discuss some of the shortcomings of current differential outcome methods. 

\textbf{Standard deviation-based measures:} 
The use of the STD in demographic group-specific error rates computed on WDI such as the EER \cite{serna_sensitive_2022}, TMR  \cite{lu_experimental_2019,krishnapriya_2019,gong_jointly_nodate,rfw}, FMR  \cite{villalobos_fair_2022,9156925,robinson_face_2020}, and FNMR  \cite{terhorst_comprehensive_2021} has been widely observed. However, by not accounting for CDIs, such approaches fail to reflect real-world scenarios where impostors may not always share demographic attributes with the genuine pairs. Additionally, quantifying biases in the form of the STD in group error rates obscures the magnitude of the bias. In other words, not using the global performance on WDIs and CDIs as a reference leads to loss of information about the significance of the bias. For instance, a model that performs three times worse than the global performance on a set of demographic groups would be treated the same as a model that performs five times worse than the global performance on the same set of groups. Depending on the fairness context, such behaviour may be undesirable. 


\textbf{Maximum disparity-based measures:}
In some metrics, bias is studied in terms of maximum discrepancy in error rates, such as the ratio of maximum and minimum values of FMRs in SER \cite{villalobos_fair_2022} or FMRs and FNMRs in IR \cite{howard_evaluating_2023} or the maximum absolute difference in FMRs and FNMRs in the FDR \cite{de2021fairness}. Quantifying bias in the form of maximum and minimum error rates obscures a model's bias and performance on demographic groups with intermediate error rates. As a result, a model biased against multiple demographic groups may be assigned the same SER, IR, or FDR score as a model biased against a single demographic group. Since all three maximum-discrepancy-based metrics quantify bias in the form of WDI performance, in practice, they suffer from the same limitation of not capturing the magnitude of the bias as the STD-based metrics because both sets of metrics lack a global reference point for measuring bias. 

\textbf{Summative aggregation-based measures:}
Mean Average Percent Error (MAPE) \cite{villalobos_fair_2022} and GARBE \cite{howard_evaluating_2023} are two metrics that quantify bias as the average of the sum of the absolute difference in group performance from a reference performance. MAPE uses a global performance consisting of CDIs and WDIs as reference, while GARBE measures how much each group's performance differs on average from all other group performances using only WDIs. MAPE accounts exclusively for FMRs, while GARBE accounts for both FMRs and FNMRs. Therefore, each metric has its own strengths and can be useful in different scenarios.

\textbf{Single error type measures:} It is common for differential outcome methods to measure the differences in FMRs \cite{villalobos_fair_2022, 9156925,robinson_face_2020}, FNMRs \cite{terhorst_comprehensive_2021}, or TMRs \cite{lu_experimental_2019,krishnapriya_2019,gong_jointly_nodate,rfw} for the different demographic groups and treat them as bias. However, neglecting the other set of errors (respectively FNMRs, FMRs, or TNMRs) can lead to missing potential biases in those errors. As a result, bias evaluation metrics that account for FMRs only, such as SER \cite{villalobos_fair_2022} and MAPE \cite{villalobos_fair_2022}, only partially capture the overall bias in a system. This means that, in theory, if a model suffers exclusively from demographic biases in the form of false non-match errors, no bias would be captured by those metrics.

\begin{table}
    \centering
    \begin{threeparttable}
        \begin{tabular}{|l|c|c|c|c|c|c|}
            \hline
             Method & C1 & C2 & C3 & L1 & L2 & L3 \\
            \hline
            STD in FMR \cite{villalobos_fair_2022,9156925,robinson_face_2020}  & \ding{51} & \ding{55} & \ding{55} & \ding{51} & \ding{51} & \ding{51}  \\
            STD in FNMR \cite{terhorst_comprehensive_2021}  & \ding{51} & \ding{55} & \ding{55} & \ding{51} & \ding{51} & \ding{51}  \\
            STD in TMR \cite{lu_experimental_2019,krishnapriya_2019,gong_jointly_nodate,rfw}  & \ding{51} & \ding{55} & \ding{55} & \ding{51} & \ding{51} & \ding{51}  \\
            STD in EER$_G$ \cite{serna_sensitive_2022} & \ding{51} & \ding{55} & \ding{55} & \ding{55} & \ding{51} & \ding{55}  \\
            IR \cite{howard_evaluating_2023} & \ding{55} & \ding{51} & \ding{55} & \ding{55} & \ding{51} & \ding{51}  \\
            FDR \cite{de2021fairness} & \ding{55} & \ding{51} & \ding{55} & \ding{55} & \ding{51} & \ding{51} \\
            SER \cite{villalobos_fair_2022} & \ding{55} & \ding{51} & \ding{55} & \ding{51} & \ding{51} & \ding{51}  \\
            MAPE \cite{villalobos_fair_2022} & \ding{55} & \ding{55} & \ding{51} & \ding{51} & \ding{55} & \ding{51}  \\
            GARBE \cite{howard_evaluating_2023} & \ding{55} & \ding{55} & \ding{51} & \ding{55} & \ding{51} & \ding{51}  \\

             SED$_G$ (Our method) & \ding{51} & \ding{55} & \ding{51} & \ding{55} & \ding{55} & \ding{55}  \\
            \hline
        \end{tabular}
        \begin{tablenotes}[flushleft]
            \item C1: Standard deviation-based measure \item C2: Maximum disparity-based measure \item C3: Summative aggregation-based measure \item L1: Accounts for a single error type \item L2: Exclusively uses WDI pairs, Lacks a global performance reference \item L3: Requires a pre-defined policy FMR \\
        \end{tablenotes}
    \end{threeparttable}
    \caption{Summary of limitations of existing differential outcome methods}
\label{tab:summary}
\end{table}
We summarize the limitations of existing differential outcome methods in Table \ref{tab:summary}. We point out that most methods, except for STD in $\mathrm{EER}_{G}$ and our proposed measure, require a policy FMR based on application or operational requirements. Hence, we see a need for a comprehensive application-independent metric that does not require any predefined operational threshold for bias quantification.

In conclusion, existing bias evaluation metrics in BV can only partially capture bias due to the limitations mentioned before. This emphasizes the need for a more comprehensive application-independent bias evaluation metric that simultaneously accounts for match and non-match errors, global CDI and WDI performance, and intermediate error rates.

\section{Demographic Bias Simulation and the Sum of Group Error Differences} \label{sec:dembias}
In this section, we describe our demographic bias simulation, synthetic dataset generation process, and our proposed measure, the Sum of Group Error Differences (SED$_G$).
\subsection{Demographic Bias Simulation}
To evaluate the demographic biases in a BV system, we typically need a dataset that consists of biometric data, demographic data, and unique person identifiers. For example, a typical bias evaluation dataset in face verification consists of face images, race labels, and unique person identifiers. To evaluate a BV system, genuine and impostor biometric sample pairs are generated randomly and sometimes selected following specific criteria, e.g. selecting only difficult pairs \cite{rfw} or alternatively using all available pairs \cite{serna_sensitive_2022}. Distances between the pairs' respective embedding vectors are then calculated to quantify different error rates, such as FMRs and FNMRs, which can be disaggregated based on the demographic membership of the individuals being represented by the biometric samples to quantify the system's demographic bias. Therefore, the minimum data required to evaluate a biometric system's bias are the distances between pairs of embedding vectors, labels indicating whether the pairs belong to the same person or not, and information about the demographic membership of the individuals being represented by the sample pairs. A biased system is then expected to give higher false matches and non-matches for some demographic groups compared to others. 

We simulate two demographic bias scenarios, one where a single group is disadvantaged at various levels in terms of model performance and a second scenario where more than one group suffer from different degrees of disadvantage in model performance. To achieve this, we synthetically generate model output distances and ground truth labels with specific FMRs at a TMR of 0.95, denoted as FMR(TMR$_{95}$), and simulate different disadvantage levels against different demographic groups using a disadvantage increase factor denoted by $x$. We describe this process more closely in the remainder of this section.




\subsubsection{Single disadvantaged group} \label{sec:single}

In the first scenario, there is a single disadvantaged group ($g_{dis}$); meanwhile, the remaining groups have the same performance levels. This scenario aims to test the bias evaluation metrics' ability to quantify disadvantage against a single demographic group. $x$ represents the disadvantage increase factor, which takes values 1, 2, 3, 5, 10, 20, and 50 depending on the level of bias being simulated. To simulate this scenario, we need to generate two synthetic datasets that satisfy the condition that the FMR at a TMR of 0.95 for the disadvantaged demographic group is $x$ times the FMR at a TMR of 0.95 for the remaining groups.

\begin{align}
 \mathrm{FMR}_{dis} (\mathrm{TMR}_{95}) &= x \cdot \mathrm{FMR}_{oth}(\mathrm{TMR}_{95}), \label{eq:FMR_dis}
\end{align}

where:
\begin{itemize}
    \item $\mathrm{FMR}_{dis}$ is the FMR at a TMR of 0.95 for the disadvantaged group $g_{dis}$.
    \item $\mathrm{FMR}_{oth}$ is the FMR at a TMR of 0.95 for the remaining groups.
    \item $x$ is the simulated disadvantage increase factor and takes the values 1, 2, 3, 5, 10, 20, and 50.
\end{itemize}

Using (\ref{eq:FMR_dis}),  we simulate different levels of demographic bias by varying the FMR at a TMR of 0.95 value of a single demographic group using the variable $x$. Using this equation, we obtain two sets of synthetic model outputs, one for the disadvantaged group denoted as D$_{dis}$ and another for all the remaining groups denoted as D$_{oth}$.

$$
\mathrm{D}_{dis} = \{d \in \mathrm{D} | \mathrm{FMR}(\mathrm{TMR}_{95}, d) = \mathrm{FMR}_{dis}(\mathrm{TMR}_{95}, d)\}
$$

$$
\mathrm{D}_{oth} = \{d \in \mathrm{D} | \mathrm{FMR}(\mathrm{TMR}_{95}, d) = \mathrm{FMR}_{oth}(\mathrm{TMR}_{95}, d)\}
$$

These two synthetic model outputs can be combined to simulate a model's output on $n$ demographic groups as follows:

$$
\mathrm{D}_{single} = \mathrm{D}_{dis} \underbrace{\cup \mathrm{D}_{oth}}_{n \text{ times}}
$$

\subsubsection{Multiple disadvantaged groups} \label{sec:multi}
In this scenario, multiple demographic groups suffer from different levels of disadvantage. This aims to test a metric's ability to capture intermediate-level biases, i.e., biases against groups other than the most disadvantaged. Multiple synthetic model outputs with varying levels of disadvantage need to be combined to simulate this scenario. For simplicity, we use the same values of $x$ as in the previous scenario and set the minimum (or best) FMR at a TMR of 0.95 denoted by FMR$_{best}$ as a reference for generating the $\mathrm{FMR} (\mathrm{TMR}_{95})$ values for the demographic groups. This process is described in (\ref{eq:FMR_i}).

\begin{align}
\mathrm{FMR}_{i} (\mathrm{TMR}_{95}) &= x \cdot \mathrm{FMR}_{best} (\mathrm{TMR}_{95}) \label{eq:FMR_i}
\end{align}

where:
\begin{itemize}
    \item $\mathrm{FMR}_{i} (\mathrm{TMR}_{95})$ is the FMR at a TMR of 0.95 of a demographic group i.
    \item $\mathrm{FMR}_{best}$ is the FMR at a TMR of 0.95 of the group with the lowest FMR at a TMR of 0.95.
    \item $x$ takes 1, 2, 3, 5, 10, 20, and 50 depending on the level of disadvantage simulated for demographic group $i$.
\end{itemize}

Using this equation, we obtain one synthetic model output dataset per each demographic group $i$ denoted by D$_{i}$.

$$
\mathrm{D}_{i} = \{d \in \mathrm{D} | \mathrm{FMR}(\mathrm{TMR}_{95}, d) = \mathrm{FMR}_{i}(\mathrm{TMR}_{95}, d)\}
$$

These outputs can then be combined to simulate the outputs of a model that is biased against more than a single group as follows:

\[ \mathrm{D}_{full} = \mathrm{D}_1 \cup \mathrm{D}_2 \cup \mathrm{D}_3 \cup ... \cup \mathrm{D}_n \]
where:
\begin{itemize}
    \item $n$ corresponds to the number of demographic groups in the simulation.
\end{itemize}

\subsubsection{Dataset generation}
To simulate the scenarios described in \ref{sec:single} and \ref{sec:multi}, we must generate model outputs with specific FMRs at a TMR of 0.95. To achieve this, we rely on hill climbing as in Algorithm \ref{alg:hillclimbing}. 

\begin{algorithm}
\caption{Hill Climbing Algorithm for BV data synthesis}
\label{alg:hillclimbing}
\begin{algorithmic}[1]
\State $\text{S\_TMR} \gets a$ \Comment{Desired TMR}
\State $\text{S\_FMR} \gets b$ \Comment{Desired FMR}
\State $\text{GT} \gets \text{array of $i$ True and $j$ False ground truth labels}$ \Comment{Binary labels for genuine and impostor pairs}
\State $\text{dist$_{gen}$} \gets \text{array of $i$ random values [0.0, 0.5]}$ \Comment{Distance values for the genuine pairs}
\State $\text{dist$_{imp}$} \gets \text{array of $j$ random values [0.5, 1.0]}$ \Comment{Distance values for the impostor pairs}
\State $\text{dist} \gets \text{concatenate dist$_{gen}$ and dist$_{imp}$}$
\State $\text{n} \gets \mathrm{\text{number of iterations}}$

\State ${\text{fitness$_{best}$}} =\left| \text{FMR}(\text{GT}, \text{dist}) - {\text{S\_FMR}} \right| + \left| \text{TMR}(\text{GT}, \text{dist}) - {\text{S\_TMR}} \right| $
\For{$i \gets 1$ to $\text{n}$} \Comment{Start the hill climbing algorithm}
    \State $\text{dist$_{new}$} \gets \text{dist + small random change}$
    \State ${\text{fitness$_{new}$}} =\left| \text{FMR}(\text{GT}, \text{dist$_{new}$}) - {\text{S\_FMR}} \right| + \left| \text{TMR}(\text{GT}, \text{dist$_{new}$}) - {\text{S\_TMR}} \right| $ 
    \If{$\text{fitness$_{new}$} < \text{fitness$_{best}$}$}
        \State $\text{dist} \gets \text{dist$_{new}$}$
        \State $\text{fitness$_{best}$} \gets \text{fitness$_{new}$}$
    \EndIf
\EndFor
\State \textbf{return} dist \Comment{Return the distance values closest to the desired values}
\end{algorithmic}
\end{algorithm}

This hill climbing algorithm performs a local search that iteratively improves a randomly generated solution, specifically fit for BV data synthesis. To this end, we use a simple fitness function that calculates the absolute difference between the desired FMR and TMR values and the actual FMR and TMR values for the input distances and ground truth labels. 
Using this algorithm, it is possible to generate different model output distances and ground truth labels that satisfy desired FMR at a TMR of 0.95 or FNMR at a TNMR of 0.95 values. In our experiment, 1000 iterations were enough to reach the desired values for our demographic bias simulation.

\subsection{Sum of Group Error Differences} \label{SED}
\textbf{SED$_G$}  is our proposed measure for simultaneously addressing the limitations of the previous metrics highlighted in Section \ref{sec:lit}. It relies on the average of the thresholds needed to achieve EER ($T_{EER_g}$) for each demographic group~($g$) in a set of demographic groups ($G$) as a reference point for measuring the deviations of individual demographic group performances from a global performance. $T_{EER_g}$ is considered as the point (threshold) where the FMR is equal to the FNMR. Hence, our proposed measure accounts for both FMRs and FNMRs and quantifies demographic bias in reference to a global performance that includes WDIs and CDIs. First, to quantify the performance (error rate) deviations, we adapt the relative difference formula as follows:

\begin{equation} \label{eq:FMR_deviation}
\delta \mathrm{FMR}_{g} = | 1 - \frac{\mathrm{FMR}_{g} (\bar{T}_{\mathrm{EER}})} {\mathrm{FMR}_{global} (\bar{T}_{\mathrm{EER}})}|
\end{equation}

\begin{equation} \label{eq:FNMR_deviation}
\delta \mathrm{FNMR}_{g} = | 1 - \frac{\mathrm{FNMR}_{g} (\bar{T}_{\mathrm{EER}})} {\mathrm{FNMR}_{global} (\bar{T}_{\mathrm{EER}})}|
\end{equation}

where:
\begin{itemize}
\item $\delta \mathrm{FMR}_{g}$ (respectively $\delta \mathrm{FNMR}_{g}$) represents the relative difference between the FMR (respectively FNMR) value for demographic group {$g$} at the average EER threshold $\bar T_{EER}$ and the global FMR (respectively FNMR) values at this same threshold. Note: We consider absolute values because we also want to treat better performance than the global reference performance as a form of bias.
\item $\mathrm{FMR}_{g}$ (respectively $\mathrm{FNMR}_{g}$)  denotes the FMR (respectively FNMR) value for demographic group $g$.
\item $\mathrm{FMR}_{global}$ (respectively $\mathrm{FNMR}_{global}$) denotes the FMR (respectively FNMR) value using the full dataset.
\end{itemize}

\begin{equation} \label{eq:SED}
\mathrm{SED}_{g} = \delta \mathrm{FMR}_{g} + \delta \mathrm{FNMR}_{g}
\end{equation}

The values of $\delta \mathrm{FMR}_{g}$ and $\delta \mathrm{FMNR}_{g}$ are summed into a single value named the Sum of Group Error Differences (SED$_g$) for simplicity, which represents the over- and under-performance of each demographic group as shown in (\ref{eq:SED}). 

\begin{equation} \label{eq:SEDG}
\mathrm{SED}_G = \{\mathrm{SED}_g \mid g \in G\}
\end{equation}

The SEDs of all demographic groups are combined in a single set $\mathrm{SED}_G$ in (\ref{eq:SEDG}). The average of the set $\mathrm{SED}_G$ represents the amount of deviation of demographic group performance from global performance. Meanwhile, the standard deviation of the set represents the variability of performance across the different demographic groups, and in that way, combines some of the strengths of aggregation- and STD-based measures.


\section{Experimental setup}
\subsection{Synthetic Dataset Description}
To test the bias evaluation metrics in the scenarios with different types and levels of biases described in Sections \ref{sec:single} and \ref{sec:multi}, model output distances and ground truth labels are created to simulate the within-demographic performance of a model with different FMR(TMR$_{95}$) values on seven hypothetical demographic groups as described in Table \ref{tab:data_char_within}. The generated distances and ground truth labels can then be combined to simulate the outputs of models with varying types and magnitudes of demographic bias.

Each set of distance and ground truth pairs represents a demographic group with 3,000 identities and two samples per identity. This allows for generating 3,000 genuine and 35,988,000 impostor pairs. However, we select an equal number of pairs (3,000 impostor and genuine pairs) to treat both types of errors with the same significance. We choose to have four demographic groups similar to some datasets in the demographic bias evaluation literature \cite{rfw}.

\begin{table}
\centering
\begin{tabular}{|c|c|c|}
\hline
\textbf{Increase factor} & \textbf{FMR$_{95}$ for g\_dis} & \textbf{Genuine:Impostor} \\
\hline
1 & 0.001 & 3000:3000 \\
\hline
2 & 0.002 & 3000:3000 \\
\hline
3 & 0.003 & 3000:3000 \\
\hline
5 & 0.005 & 3000:3000 \\
\hline
10 & 0.01 & 3000:3000 \\
\hline
20 & 0.02 & 3000:3000 \\
\hline
50 & 0.05 & 3000:3000 \\
\hline
\end{tabular}
\caption{Within-demographic only synthetic dataset characteristics}
\label{tab:data_char_within}
\end{table}

\begin{table}
\centering
\begin{tabular}{|c|c|c|}
\hline
\textbf{Increase factor} & \textbf{FMR$_{95}$} & \textbf{Genuine:Impostor} \\
\hline
0.1 & 0.0001 & 3000:24000 \\
\hline
\end{tabular}
\caption{Within and cross-demographic synthetic dataset characteristics}
\label{tab:data_char_cross}
\end{table}

For simplicity, we refer to FMR(TMR$_{95}$) as FMR$_{95}$ in the remainder of the paper. 

An additional set of simulated model output distances and ground truth labels is also created to simulate a model's global performance as described in Table \ref{tab:data_char_cross}. This performance is usually better than within-demographic performance because it is dominated by cross-demographic impostor pairs, which are the easiest for a typical BV system to distinguish. Therefore, the number of genuine pairs becomes 12,000, and while it is possible to generate 575,952,000 impostor pairs, 600,000 impostor pairs are enough to simulate a global performance with an FMR$_{95}$ of 0.0001.

\subsection{Compared Bias Evaluation Metrics} 
We test our proposed dual metrics and other established metrics, such as IR and GARBE \cite{villalobos_fair_2022}, FDR \cite{de2021fairness}, and STD in EER$_G$ \cite{serna_sensitive_2022}. To allow an equitable comparison, we only use metrics that account for both FMRs and FNMRs to quantify the bias in our scenarios with different levels and types of bias. The evaluated metrics are described in detail in the following.

\textbf{IR} is a maximum-disparity-based metric that combines the ratio between the highest and lowest FMR and FNMR values among a set of demographic groups. It calculates the maximum disparity in FMRs and FNMRs across different demographic groups at a desired or policy FMR threshold ($T_{FMR_p}$).

\begin{equation}
\mathrm{A(T_{FMR_p})} = \frac{{\max \left\{ \mathrm{FMR}_{g}(T_{\mathrm{FMR}_p}), \forall g \in G \right\}}}{{\min \left\{ \mathrm{FMR}_{g}(T_{\mathrm{FMR}_p}), \forall g \in G \right\}}}
\end{equation}
\begin{equation}
\mathrm{B(T_{FMR_p}}) = \frac{{\max \left\{ \mathrm{FNMR}_{g}(T_{\mathrm{FNMR}_p}), \forall g \in G \right\}}}{{\min \left\{ \mathrm{FNMR}_{g}(T_{\mathrm{FNMR}_p}), \forall g \in G \right\}}}
\end{equation}
\begin{equation}
\mathrm{IR} = A(T)^\alpha B(T)^{1-\alpha}
\end{equation}
where:
\begin{itemize}
    \item $\alpha$ is the hyperparameter that determines the weight assigned to FMRs relative to FNMRs during the calculation of FDR and can be adjusted according to operational requirements.
\end{itemize}
We set the alpha ($\alpha$) hyperparameter for all of the metrics to 0.5 to treat FMRs and FNMRs with the same level of significance so that they are comparable.\newline

\textbf{FDR} is also a maximum-disparity-based metric that quantifies the rate of false match and non-match errors. Similar to IR, FDR takes into account the maximum disparity (in the form of a maximum difference rather than a ratio) in FMRs and, additionally, FNMRs at $T_{FMR_p}$. 

\begin{equation} \label{eq:FDR}
\begin{split}
\mathrm{FDR} = 1 - \big[&\alpha (\max \{ | \mathrm{FMR}_{g1}(T_{\mathrm{FMR}_p})  \\
&\quad - \mathrm{FMR}_{g2}(T_{\mathrm{FMR}_p}) |, \forall g1,g2 \in G \} ) \\
&\quad + (1 - \alpha) (\max \{ | \mathrm{FNMR}_{g1}(T_{\mathrm{FMR}_p}) \\ 
&\quad - | \mathrm{FNMR}_{g2}(T_{\mathrm{FMR}_p}) |, \forall g1,g2 \in G \} ) \big]
\end{split}
\end{equation}

\textbf{GARBE} is a metric inspired by a measure of statistical dispersion called the Gini coefficient, used to quantify the average difference between the FMRs and FNMRs of available demographic groups at a predefined policy FMR threshold.
\begin{equation} \label{eq:MAPE}
\begin{split}
\mathrm{Gini}_x(T_{\mathrm{FMR}_p}) = \big(\frac{|G|}{|G| - 1}\big) \big(\frac{| x_{g1}(T_{\mathrm{FMR}_p}) - x_{g2}(T_{\mathrm{FMR}_p}) |}{2n^2\bar{x}}\big), \\
\forall g1,g2 \in G\
\end{split}
\end{equation}

\begin{equation}
\mathrm{A} = \mathrm{Gini}_{\mathrm{FMR}}(T_{\mathrm{FMR}_p}); \mathrm{B} = \mathrm{Gini}_{\mathrm{FNMR}}(T_{\mathrm{FMR}_p}) 
\end{equation}

\begin{equation}
 GARBE = \alpha A + (1 - \alpha)B    
\end{equation}



\textbf{STD in EER$_G$ ($\sigma_{\mathrm{EER}_G}$)}  is a simplistic measure of bias that quantifies the standard deviation in demographic group-specific EERs computed on within-demographic pairs (see (\ref{eq:STDinEER})), with EER being the value at which the FMR is equal to the FNMR (see (\ref{eq:EER})).

\begin{equation} \label{eq:EER}
\mathrm{EER}_g = \frac{\mathrm{FMR}_g + \mathrm{FNMR}_g}{2} \ \big| \ \mathrm{FMR}_g = \mathrm{FNMR}_g 
\end{equation}

\begin{equation} \label{eq:EER_G}
\mathrm{EER}_G = \{\mathrm{EER_g} \mid g \in G\}
\end{equation}
$\mathrm{EER}_G$ represents the set of all demographic group-specific EER values, with each $\mathrm{EER}_g$ corresponding to a specific demographic group $g$ in the set of all demographic groups $G$. Note: In \cite{serna_sensitive_2022}, the EER is expressed in percentage form.

\begin{equation} \label{eq:STDinEER}
 \sigma_{\mathrm{EER}_G} = \sqrt{\frac{1}{|\mathrm{EER}_G|}\sum_{\mathrm{EER}_g \in \mathrm{EER}_G}(\mathrm{EER}_g - \mu_{\mathrm{EER}_G})^2}
\end{equation}

\textbf{STD in SED$_G$ and Average of SED$_G$}  are the proposed metrics in our dual-metric measure, designed to simultaneously address the limitations of the previous metrics highlighted in Section \ref{sec:lit}. Set SED$_G$ is described in detail in Section \ref{SED}. The average of set SED$_G$ values ($\overline{\mathrm{SED}}_G$) measures the magnitude of the deviation of demographic group performance from global performance, while the STD in set SED$_G$ ($\sigma_{\mathrm{SED}_G}$) measures the degree of variation in demographic group performance.

\section{Experimental Results and Discussion}
The experiments aim to test the efficacy of the bias evaluation metrics in quantifying the biases of BV systems with different types and magnitudes of bias using the scenarios described in Sections \ref{sec:single} and \ref{sec:multi}.
\subsection{Single Disadvantaged Group} \label{sec:one} 
\begin{table} \caption{Evaluation metrics at different FMR$_{95}$ values for a single disadvantaged group}
\label{tab:singdis}
\begin{tabular}{|c|c|c|c|c|c|c|c|}
\hline
Ratios & IR & GARBE & FDR & $\sigma_{\mathrm{EER_G}}$ & $\sigma_{\mathrm{SED_G}}$ & $\overline{\mathrm{SED}}_G$ \\ \hline
1:1:1:1     & 1.0  & 0.0000 & 1.00   & 0.00          & 0.00   & 0.24 \\ \hline 
1:1:1:2  &   1.33  & 0.0090  & 0.9990  & 8.66e-4  &  0.17 & 0.32  \\ \hline
1:1:1:3   &  3.63  & 0.0397  & 0.9958  & 1.29e-3 &  0.28 & 0.85  \\ \hline
1:1:1:5     &  13.62 &  0.0758  & 0.9923  & 3.96e-3  &  0.75 & 1.26 \\ \hline
1:1:1:10    &  22.40  &  0.0891  & 0.9890  & 4.69e-3  & 0.91 & 1.30 \\ \hline
1:1:1:20    &  87.54  &  0.1170  & 0.9821  & 6.92e-3  &  1.32 & 1.56 \\ \hline
1:1:1:50     &  368.7  & 0.1427  & 0.9693  & 1.54e-2  &  2.99 & 2.55 \\ \hline
\end{tabular}
\end{table}

In the case of a single disadvantaged group $g_{dis}$ (4th group in the ratios in Table \ref{tab:singdis}), maximum disparity metrics such as IR and FDR rank all systems accurately due to the absence of intermediate-level biases. All metrics consistently report higher scores for systems with a greater FMR$_{95}$ for the disadvantaged group, making them equally usable in this scenario. It is worth mentioning that for system 1:1:1:1, four metrics, IR, GARBE, FDR, STD in EER$_G$, and STD in SED$_G$, report no biases in their score. This is because such metrics measure the bias based on the maximum and minimum error difference or the variation in demographic group errors. The Average of SED$_G$ is the only metric that indicates a bias score equal to 0.24, as it measures bias by comparing the errors for each demographic group against the global errors. Depending on the fairness context, any of the metrics can be suitable for this scenario.

\subsection{Multiple Disadvantaged Groups}
\subsubsection{Two-disadvantaged groups (Table \ref{table:two-dis})} \label{sec:two}
\begin{table}
\caption{Evaluation metrics at different FMRs at a TMR of 0.95 ratios for two disadvantaged groups} \label{table:two-dis}
\centering
\begin{tabular}{|c|c|c|c|c|c|c|}
\hline
Ratios & IR & GARBE & FDR & $\sigma_{\mathrm{EER_G}}$ & $\sigma_{\mathrm{SED_G}}$ & $\overline{\mathrm{SED}}_G$ \\
\hline 
\textcolor{purple}{1:1:2:2} & 1.33 & 0.0120 & 0.9990 & 9.99e-4 & 0.20 & 0.37 \\
\hline 
\textcolor{green!50!black}{1:1:2:3} & 3.63 & 0.0428 & 0.9958 & 1.78e-3 & 0.35 & 0.81 \\
\hline 
\textcolor{green!50!black}{1:1:2:5} & 13.62 & 0.0793 & 0.9923 & 4.33e-3 & 0.82 & 1.15 \\
\hline 
\textcolor{purple}{1:1:3:3} & 3.63 & 0.0530 & 0.9958 & 1.50e-3 & 0.38 & 1.05 \\
\hline 
\textcolor{green!50!black}{1:1:3:5} & 13.62 & 0.0905 & 0.9923 & 3.74e-3 & 0.70 & 1.32 \\
\hline 
\textcolor{purple}{1:1:5:5} & 13.62 & 0.1011 & 0.9923 & 4.58e-3 & 0.86 & 1.68 \\
\hline
\end{tabular}
\end{table}

\begin{itemize}
    \item IR is a maximum-disparity-based metric, meaning it relies on maximum and minimum FMR values. Therefore, it fails to capture intermediate-level FMR$_{95}$ value changes. Hence it assigns identical scores to \textcolor{green!50!black}{1:1:2:3} and \textcolor{purple}{1:1:3:3} although system \textcolor{purple}{1:1:3:3} is simulated to have a higher disadvantage for the third group. A similar case is observed for \textcolor{green!50!black}{1:1:2:5}, \textcolor{green!50!black}{1:1:3:5}, and \textcolor{purple}{1:1:5:5}. Similarly, FDR is a maximum disparity-based metric that relies on minimum and maximum FMR and FNMR values. Therefore, it also fails to capture intermediate-level biases for those same examples.
    \item STD in EER$_G$ and STD in SED$_G$ give a lower bias score to \textcolor{green!50!black}{1:1:3:5} in comparison to \textcolor{green!50!black}{1:1:2:5} although the former is simulated to have a higher disadvantage for the third group. This is because they are both STD-based measures and in that sense, system \textcolor{green!50!black}{1:1:3:5} has a lower degree of variation in group errors. The only metrics capable of correctly quantifying the magnitude of bias, in this case, are GARBE and Average of SED$_G$.
    \item The only metrics with no failure cases are GARBE and the Average of SED$_G$ followed by STD in SED$_G$ and STD in EER$_G$, which both share the same failure cases due to their shared STD-based characteristic.
\end{itemize}

\subsubsection{Three-disadvantaged groups (Table \ref{tab:three-dis})} \label{sec:three}

\begin{table} 
\caption{Evaluation metrics at different FMR$_{95}$ ratios for three disadvantaged groups} \label{tab:three-dis}
\begin{tabular}{|c|c|c|c|c|c|c|}
\hline
Ratios & IR & GARBE & FDR & $\sigma_{\mathrm{EER_G}}$ & $\sigma_{\mathrm{SED_G}}$ & $\overline{\mathrm{SED}}_G$ \\
\hline 
\textcolor{purple}{1:2:2:2} & 1.33 & 0.0090 & 0.9990 & 8.66e-4 & 0.16 & 0.46 \\
\hline 
1:2:2:3 & 3.63 & 0.0399 & 0.9958 & 0.0399 & 0.39 & 0.69 \\
\hline 
1:2:2:5 & 13.62 & 0.0767 & 0.9920 & 0.0767 & 0.85 & 0.98 \\
\hline 
\textcolor{green!50!black}{1:3:3:2} & 3.63 & 0.0502 & 0.9958 & 2.12e-3 & 0.43 & 0.94 \\
\hline 
\textcolor{purple}{1:3:3:3} & 3.63 & \textcolor{red}{0.0397} & 0.9958 & \textcolor{red}{1.29e-3} & \textcolor{red}{0.37} & 1.23 \\
\hline 
1:3:3:5 & 13.62 & 0.0786 & 0.9923 & 3.33e-3 & 0.58 & 1.41 \\
\hline 
\textcolor{green!50!black}{1:5:5:2} & 13.62 & 0.0990 & 0.9923 & 5.13e-3 & 0.95 & 1.50 \\
\hline 
\textcolor{green!50!black}{1:5:5:3} & 13.62 & \textcolor{red}{0.0906} & 0.9923 & \textcolor{red}{3.97e-3} & \textcolor{red}{0.73} & 1.70 \\
\hline 
\textcolor{purple}{1:5:5:5} & 13.62 & \textcolor{red}{0.0758} & 0.9923 & \textcolor{red}{3.96e-3} & \textcolor{red}{0.73} & 2.02 \\
\hline
\end{tabular}
\end{table}

\begin{itemize}
    \item The highlighted cases in red help demonstrate the functionalities of GARBE, STD in EER$_G$, and STD in SED$_G$. All three metrics quantify the difference in performance between demographic groups rather than the bias in each group separately against a reference point. Depending on the fairness context and definition, it is likely that system \textcolor{green!50!black}{1:5:5:2} is considered less biased than systems \textcolor{green!50!black}{1:5:5:3} and \textcolor{purple}{1:5:5:5}, which suffer from a higher level of disadvantage for the fourth group. However, the three metrics fail to capture this relation and, therefore, might not be suitable for this scenario.
    \item Similar to the previous scenario, the scores of IR and FDR are dictated by the highest FMR$_{95}$, causing them to neglect intermediate biases. This is also observable when comparing elements from Tables \ref{table:two-dis} and \ref{tab:three-dis}. Systems \textcolor{purple}{1:1:2:2} and \textcolor{purple}{1:2:2:2} are given the same IR and FDR scores, 1.33 and 0.9990, respectively. Similarly, systems \textcolor{green!50!black}{1:1:3:2}, \textcolor{purple}{1:1:3:3}, \textcolor{green!50!black}{1:3:3:2}, and \textcolor{purple}{1:3:3:3} all have the same IR and FDR scores, 3.63 and 0.9958 respectively. A distinction between such systems might be necessary depending on the fairness definition.
    \item The dual SED$_G$ metrics provide a good understanding of the type and magnitude of bias in each system. The Average of SED$_G$ helps us understand the overall bias present in the system, while the STD in SED$_G$ helps us understand the difference in performance across demographic groups. When used together, these metrics provide the most accurate results in this scenario.
\end{itemize}

\subsubsection{Four-disadvantaged groups (Table \ref{tab:four-dis})} \label{sec:four}
\begin{table}
\caption{Evaluation metrics at different FMR$_{95}$ ratios for four disadvantaged groups} \label{tab:four-dis}
\begin{tabular}{|c|c|c|c|c|c|c|}
\hline
Ratios & IR & GARBE & FDR & $\sigma_{\mathrm{EER_G}}$ & $\sigma_{\mathrm{SED_G}}$ & $\overline{\mathrm{SED}}_G$ \\
\hline 
\textcolor{purple}{2:2:2:2} & 1.00 & 0.0000 & 1.00 & 0.00 & 0.00 & 0.49 \\
\hline 
2:2:2:3 & 2.72 & 0.0310 & 0.9968 & 2.16e-3 & 0.44 & 0.70 \\
\hline 
2:2:2:5 & 10.20 & 0.0682 & 0.9933 & 4.83e-3 & 0.92 & 0.93\\
\hline 
2:2:3:3 & 2.72 & 0.0413 & 0.9968 & 2.50e-3 & 0.46 & 0.97 \\
\hline 
2:2:3:5 & 10.20 & 0.0795 & 0.9933 & 4.59e-3 & 0.83 & 1.24\\
\hline 
2:2:5:5 & 10.20 & 0.0909 & 0.9933 & 5.58e-3 & 1.07 & 1.46 \\
\hline 
2:3:3:3 & 2.72 & 0.0310 & 0.9968 & 2.16e-3 & 0.48 & 1.29 \\
\hline 
2:3:3:5 & 10.20 & 0.0702 & 0.9933 & 3.95e-3 & 0.65 & 1.43 \\
\hline 
2:3:5:5 & 10.20 & 0.0826 & 0.9933 & 4.68e-3 & 0.82 & 1.69 \\
\hline 
2:5:5:5 & 10.20 & 0.0682 & 0.9933 & 4.83e-3 & 0.92 & 2.00 \\
\hline 
\textcolor{purple}{3:3:3:3} & 1.00 & 0.0000 & 1.00 & 0.00 & 0.00 & 1.77 \\
\hline 
3:3:3:5 & 3.74 & 0.0402 & 0.9965 & 2.67e-3 & 0.35 & 1.89 \\
\hline 
3:3:5:5 & 3.74 & 0.0536 & 0.9965 & 3.08e-3 & 0.45 & 1.91 \\
\hline 
3:5:5:5 & 3.74 & 0.0402 & 0.9965 & 
2.67e-3 & 0.41 & 2.19 \\
\hline 
\textcolor{purple}{5:5:5:5} & 1.00 & 0.0000   & 1.00 & 0.00 & 0.00 & 2.53 \\
\hline
\end{tabular}
\end{table}
In this scenario, all four groups suffer from some level of disadvantage in reference to global performance.
\begin{itemize}
    \item Only the Average of SED$_G$ metric is capable of differentiating between systems \textcolor{purple}{2:2:2:2}, \textcolor{purple}{3:3:3:3}, and \textcolor{purple}{5:5:5:5}. This makes it useful in all four scenarios discussed so far, having the advantage of providing additional information about the type of bias present when used alongside the STD in SED$_G$. This example shows the benefit of using a metric that quantifies bias in terms of a difference from a global performance because it can capture the magnitude of the disadvantage for all the groups, as opposed to metrics that consider deviations and maximum differences in group error rates as bias. 
    \item  When all four groups have the same level of disadvantage as in systems \textcolor{purple}{2:2:2:2}, \textcolor{purple}{3:3:3:3}, \textcolor{purple}{5:5:5:5}, WDI-based metrics that lack a global reference such as IR, GARBE, FDR, STD in EER$_G$, and STD in SED$_G$ cannot quantify the magnitude of this disadvantage and assigns them the same score. As previously discussed, this behaviour, in some cases, may be undesirable.
    \item Similar to the scenarios discussed in Sections \ref{sec:two} and \ref{sec:three}, IR and FDR fail to capture intermediate-level biases for all systems.  Additionally, for systems \textcolor{purple}{2:2:2:2}, \textcolor{purple}{3:3:3:3}, and \textcolor{purple}{5:5:5:5} which do not suffer intermediate biases but exhibit the same level of disadvantage for all four groups, both metrics assign them identical scores. As a result, IR and FDR are completely unusable when all demographic groups suffer from some level of disadvantage i.e. this scenario.
    \item As previously stated, when the Average of SED$_G$ is used alongside the STD in SED$_G$, it is possible to measure the magnitude of bias present and also get an idea about the type of bias present. For instance, STD in SED$_G$ assigns the same score (a value of zero) to systems \textcolor{purple}{2:2:2:2}, \textcolor{purple}{3:3:3:3}, and \textcolor{purple}{5:5:5:5} that have the same type of bias. This enables us to know that the level of bias is consistent among all the groups, and by examining the Average of SED$_G$, it is possible to know the magnitude of this bias.
\end{itemize}

\subsection{Bias in False Non-match Errors} \label{sec:biasfme}
We conduct an additional experiment to evaluate the effectiveness of the bias evaluation metrics in capturing bias in the form of false non-match errors. In this experiment, we fixed the FMR value to 0.05, which is equivalent to a TNMR of 0.95. We then introduce the bias by varying the values of FNMR at a TNMR of 0.95 for a single disadvantaged group. We only simulate a single disadvantaged group scenario as the purpose of this experiment is to test the metrics' ability to detect bias in the form of false non-match errors. We tested and demonstrated the limitations of the metrics in the multiple disadvantaged group scenarios in the previous experiments. 
We use the notation FNMR$_{95}$ to represent the FNMR at a TNMR of 0.95, similar to how FMR$_{95}$ represents the FMR at a TMR of 0.95.

\begin{table} \caption{Evaluation metrics at different FNMR$_{95}$ values for a single disadvantaged group}
\label{tab:dis}
\begin{tabular}{|c|c|c|c|c|c|c|c|}
\hline
Ratios & IR & GARBE & FDR & $\sigma_{\mathrm{EER_G}}$ & $\sigma_{\mathrm{SED_G}}$ & $\overline{\mathrm{SED}}_G$ \\ \hline
1:1:1:1     & 1  & 0.0000 & 1.00   & 0.00          & 0.00   & 2.14 \\ \hline 
1:1:1:2  &   13  & 0.0745  & 0.9963  & 2.02e-3  &  1.63 & 3.07  \\ \hline
1:1:1:3   &  64  & 0.1166  & 0.9913  & 3.60e-3 &  2.76 & 4.04  \\ \hline
1:1:1:5     &  134 &  0.1263  & 0.9890  & 5.19e-3  &  3.42 & 4.55 \\ \hline
1:1:1:10    &  502 &  0.1572  & 0.9821  & 7.93e-3  & 6.47 & 6.18 \\ \hline
1:1:1:20    &  1136  &  0.1667  & 0.977  & 1.04e-2  &  6.96 & 7.48 \\ \hline
1:1:1:50     &  9396  & 0.1981  & 0.9568  & 1.94e-2  &  14.99 & 10.69 \\ \hline
\end{tabular}
\end{table}
The behaviour of the metrics in this scenario (Table \ref{tab:dis}) is identical to their behaviour in the scenario where bias is introduced in the form of different FMR$_{95}$ values for a single disadvantaged group in Section \ref{sec:one}. All metrics are capable of accurately ranking the systems, which is because all the tested metrics rely on false match and non-match errors for bias quantification. The Average of SED$_G$ is the only metric scoring bias for system 1:1:1:1, which might be useful depending on the fairness definition. IR and FDR rank all systems accurately, mainly due to the absence of intermediate-level biases.

\section{Conclusion}
In conclusion, we revisit the bias evaluation metrics and discuss some of their advantages, disadvantages, and nuances.

\textbf{IR and FDR}: Since IR and FDR are both based on maximum disparity, it is sensible to discuss them together. Additionally, they rank all systems similarly. They are most suitable when there is only a single disadvantaged group. Being based on maximum disparity, both metrics study biases in terms of maximum and minimum error rates, hence not capturing intermediate biases. They are unusable in the presence of intermediate biases or when all demographic groups exhibit some level of disadvantage.

\textbf{GARBE}: GARBE quantifies bias in terms of the difference in false match and non-match errors among the demographic groups. This makes it suffer from similar issues and failure cases as the STD-based metrics, STD in EER$_G$ and STD in SED$_G$. Since it relies exclusively on WDIs, it does not capture the magnitude of bias making it unusable in scenarios where all demographic groups suffer from the same level of disadvantage.

\textbf{STD in EER$_G$}: STD in EER$_G$ is a WDI-based metric that considers both false match and non-match errors and measures the variation in demographic group error rates. While this means that it captures intermediate biases, it also means that it does not provide information about the magnitude of the bias present, which is an advantage that Average of SED$_G$ has when used alongside STD in SED$_G$.

\textbf{STD in SED$_G$ and Average of SED$_G$}: the value of the SED by design takes into account the differences in false match and false non-match errors in reference to the global values of those errors (see (\ref{eq:SED})). By accounting for global CDI and WDI performance, SED$_G$ metrics hold a clear advantage over STD in EER$_{G}$. In our simulation, STD in EER$_{G}$ and STD in SED$_{G}$ ranked systems similarly. However, only the Average of SED$_{G}$ metric ranked all systems correctly. In combination, the SED$_G$ metrics provide a clearer understanding of the magnitude and type of bias present in a system making them usable in all scenarios.

In summary, we present our proposed measure of demographic bias in BV as a dual-metric measure that overcomes the identified limitations of previous metrics. In combination, the dual metrics in our measure offer an understanding about the type and magnitude of bias present in a BV system. Therefore, we encourage the research community around BV systems to include them in their evaluation.
\section{Acknowledgments}
This work has received funding from the European Union’s Horizon 2020 research and innovation programme under Marie Sklodowska-Curie Actions (grant agreement number 860630) for the project ‘‘\href{https://nobias-project.eu}{NoBIAS - Artificial Intelligence without Bias}’’. This work reflects only the authors’ views and the European Research Executive Agency (REA) is not responsible for any use that may be made of the information it contains.

\balance
{\small
\bibliographystyle{ieee}
\bibliography{main}
}

\end{document}